\documentclass[10pt,twocolumn,letterpaper]{article}

\usepackage{times}
\usepackage{epsfig}
\usepackage{graphicx}
\usepackage{amsmath}
\usepackage{amssymb}
\usepackage{multirow}
\usepackage{graphicx}
\usepackage{subfigure}
\usepackage{array}
\usepackage{color}
\usepackage{comment}
\usepackage{booktabs}
\usepackage{threeparttable}

\usepackage[pagebackref=true,breaklinks=true,letterpaper=true,colorlinks,bookmarks=false,colorlinks,linkcolor=red]{hyperref}

\begin{document}

\title{APE-GAN: Adversarial Perturbation Elimination with GAN}

\author{
	Shiwei Shen\\
ICT\footnotemark[1]\\
{\tt\small shenshiwei@ict.ac.cn}
\and
Guoqing Jin\\
ICT\footnotemark[1]\\
{\tt\small jinguoqing@ict.ac.cn}
\and
Ke Gao\\
ICT\footnotemark[1]\\
{\tt\small kegao@ict.ac.cn}
\and
Yongdong Zhang\\
ICT\footnotemark[1]\\
{\tt\small zhyd@ict.ac.cn}
}
\date{}

\maketitle

\renewcommand{\thefootnote}{\fnsymbol{footnote}}
\footnotetext[1]{Institute of Computing Technology
	Chinese Academy of Sciences,Beijing 100190, China}

\begin{abstract}
	Although neural networks could achieve state-of-the-art performance while
	recongnizing images, they often suffer a tremendous defeat from adversarial examples--inputs generated by utilizing imperceptible but intentional perturbation to clean samples from the datasets. How to defense against adversarial examples is an important problem which is well worth researching. So far, very few methods have provided a significant defense to adversarial examples. 
	In this paper, a novel idea is proposed and an effective framework based Generative Adversarial Nets named APE-GAN is implemented to defense against the adversarial examples.
	The experimental results on three benchmark datasets including MNIST, CIFAR10 and ImageNet indicate that APE-GAN is effective to resist adversarial examples generated from five attacks.
	
\end{abstract}

\section{Introduction}

Deep neural networks have recently achieved excellent performance on a variety of visual and speech recognition tasks.
However, they have non-intuitive characterisitics and intrinsic blind spots that are easy to attack using obscure manipulation of their input\cite{goodfellow2014explaining,kurakin2016adversarial,papernot2016transferability,szegedy2013intriguing}.
In many cases, the structure of the neural networks is strongly related to the training data distribution, which is in contradiction with the network's ability to achieve high generalization performance.

Szegedy et al. \cite{szegedy2013intriguing} first noticed that imperceptible perturbation of test samples can be misclassified by neural networks.
They term this kind of subtly perturbed samples \textquotedblleft \textit{adversarial examples}\textquotedblright.
In contrast to noise samples, adversarial examples are imperceptible, designed  intentionally, and more likely to cause false predictions in the image
classification domain.
What is more serious is that adversarial examples transfer across models named \emph{transferability}, which can be leveraged to perform \emph{black-box attacks}\cite{liu2016delving,papernot2016transferability}.
In other words, an adversary can find the adversarial examples generated from \emph{substitute model} trained by the adversary and apply it to attack the \emph{target model}. Howerver, so far, transferability is mostly appears on small datasets, such as MNIST and CIFAR10. Transferability over large scale datasets, such as ImageNet, has yet to be better understood. Therefore, in this paper, black-box attacks is not taken into consideration, and resisting the \emph{white-box attacks} (\textbf{the adversary has complete access to the target model including the architecture and all paramaters}) is the core work of the paper.



\begin{figure}
	\centering
	\includegraphics[width=0.99\linewidth]{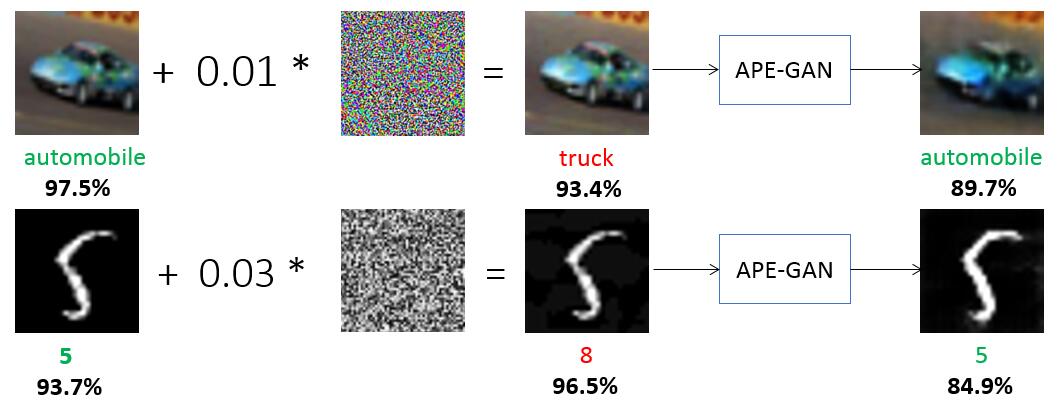}
	\caption{A demonstration of APE-GAN applied to MNIST and CIFAR10 classification networks. The images with a small perturbation named adversarial samples cannot be correctly classified with high confidence. However, the samples processed by our model is able to be classified correctly.}
	\label{fig:demonstration}
\end{figure}

Adversarial examples pose potential security threats for practical machine learning applications. Recent research has shown that a large fraction of adversarial examples are classified incorrectly even when obtained from the cell-phone camera\cite{kurakin2016adversarial}.
This makes it possible that an adversary crafts adversarial images of traffic signs to cause the self-driving cars to take unwanted actions\cite{papernot2016practical}. Therefore, the research of resisting adversarial examples is very significant and urgent.

Until now, there are two classes of approaches to defend against adversarial examples. The straightforward way is to make the model inherently more robust with enhanced training data as show in Figure \ref{fig:generator} or adjusted learning strategies. \emph{Adversarial training} \cite{goodfellow2014explaining,tramer2017ensemble} or \emph{defensive distillation} \cite{papernot2017extending,papernot2016distillation} belongs to this class.
It is noteworthy that the original defensive distillation is broken by Carlini and Wagner's attack \cite{carlini2017towards}, however, which can be resisted by the extending defensive distillation. In addition, the faultiness original 
adversarial training remains highly vulnerable to \emph{transferred} adversarial examples crafted on other models which is discussed in the \emph{ensemble adversarial training},
and the model trained using ensemble adversarial training are slightly less robust to some white-box attacks.
The second is a series of detection mechanisms used to detect and reject an adversarial sample\cite{gong2017adversarial,metzen2017detecting}. 
Unfortunately, Carlini et al. \cite{carlini2017adversarial} indicate that adversarial examples generated from Carlini and Wagner's attack are significantly harder to detect than previously appreciated via bypassing ten detection methods. Therefore, defensing
against adversarial examples is still a huge challenge.

Misclassification of the adversarial examples is mainly
due to the intentionally imperceptible perturbations to
some pixels of the input images. Thus, we propose an algorithm
to eliminate the adversarial perturbation of input data to defense
against the adversarial examples. The adversarial perturbation elimination
of the samples can be defined as the problem of learning
a manifold mapping from adversarial examples to original
examples. Generative Adversarial Net (GAN) proposed by
Goodfellow et al\cite{goodfellow2014generative} is able to generate
images similar to training set with a random noise. Therefore,
we designed an framework utilizing GAN to generate clean examples from adversarial examples. Meanwhile, SRGAN\cite{ledig2016photo}, the successful application of GAN
on super-resolution issues, provides a valuable experience to
the implementation of our algorithm.

%


In this paper, an effective framework is implementated to
eliminate the aggressivity of adversarial examples before being
recognized, as shown in Figure \ref{fig:generator}. 
The code and trained models of the framework are available on \url{https://github.com/shenqixiaojiang/APE-GAN}, so we welcome new attacks to break our defense.



This paper makes the following \textbf{contributions}:

\begin{figure*}
	\centering
	\includegraphics[width=1\linewidth]{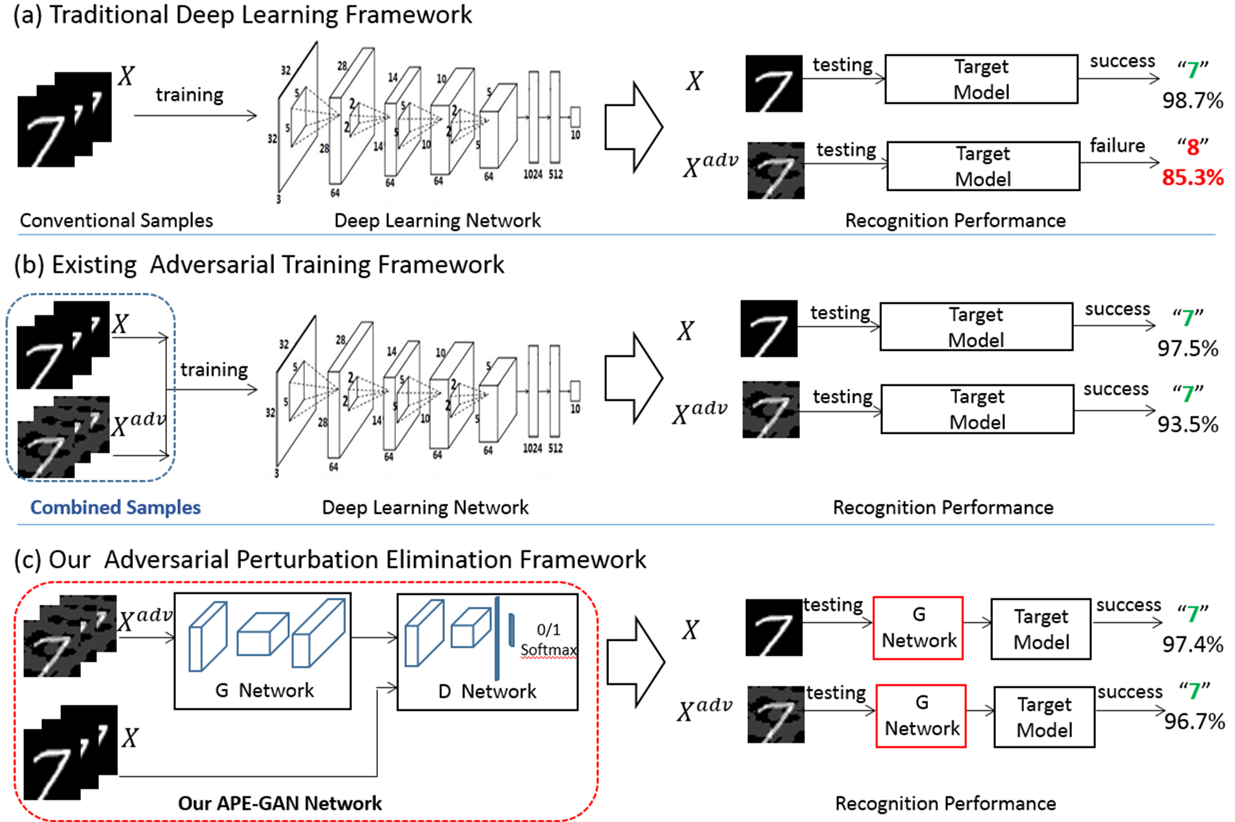}
	\caption{ (a) The traditional deep learning framework shows its robustness to clean images but is highly vulnerable to adversarial examples. (b) Existing adversarial training framework can increase the robustness of target model with enhanced training data. (c) We propose an adversarial perturbation elimination framework named APE-GAN to eliminate the perturbation of the adversarial examples before feeding it into the target model to increase the robustness.}
	\label{fig:generator}
\end{figure*}

\begin{itemize}
	
	\item \textbf{ A new perspective of defending against adversarial examples is proposed.} The idea is to first eliminate the adversarial perturbation using a trained network and then feed the processed example to classification networks. 
	
	\item \textbf{ An effective and reasonable framework based on the above idea is implemented to resist adversarial examples.} The experimental results on three benchmark datasets demonstrate the effectiveness.

	\item \textbf{ The proposed framework possesses strong applicability.} It can tackle adversarial examples without knowing what target model they are constructed upon. 
	
	\item \textbf{The training procedure of the APE-GAN
		needs no knowledge of the architecture and parameters of the target model.}
	
\end{itemize}








\section{Related Work}
In this section, methods of generating adversarial examples that are closely related to this work is briefly reviewed. 
In addition, GAN and its connection to our method will be discussed.


In the remaining of the paper we use the following notation and terminology:
\begin{itemize}
	\item $X$ - the clean image from the datasets.
	\item $X^{adv}$ - the adversarial image.
	\item $y_{true}$ - the true class for the input $X$.
	\item $y_{fool}$ - the class label different from $y_{true}$ for the input $X$.
	\item $f$ - the classifier mapping from input image to a discrete label set.
	\item $J(X,y)$ - the cost function used to train the model given image $X$ and class $y$.
	\item $\epsilon$ - the size of worst-case perturbations, $\epsilon$ is the upper bound of  the $L_{\infty}$ norm of the perturbation.
	\item $Clip_{X,\epsilon}\{A\}$ - an $\epsilon$-neighbourhood clipping of $A$ to the range $[X_{i,j}-\epsilon,X_{i,j}+\epsilon]$.
	\item Non-targeted adversarial attack - the goal of it is to slightly modify clean image that it will be classified incorrectly by classifier.
	\item Targeted adversarial attack - the goal of it is to slightly modify source image that it will be classified as specified target class by classifier.
\end{itemize}



\subsection{Methods Generating Adversarial Examples}

In this subsection, six approaches we utilized to generate adversarial images are provided with a brief description.

\subsubsection{L-BFGS Attack}
Minimum distortion generation function $D$ where $f(D(x))\neq f(x)$ defined as an optimization problem can be solved by a box-constrained L-BFGS to craft adversarial perturbation under $L_{2}$ distance metric \cite{szegedy2013intriguing}.

The optimization problem can be formulized as:

\begin{eqnarray}\label{eq:L_BFGS_adv}
	\begin{aligned}
		&minimize \quad \lambda \cdot ||D(x) - x||_2^2 +  J(D(x),y_{fool}) \\
		&subject \ to  \quad D(x) \in [0,1]
	\end{aligned}
\end{eqnarray}

The constant $\lambda > 0$ controls the trade-off between the perturbation's amplitude and its attack power which can be found with line search.

\subsubsection{Fast Gradient Sign Method Attack (FGSM)}
Goodfellow et al.\cite{goodfellow2014explaining} proposed this method to generate adversarial images under $L_{\infty}$ distance metric. 

Given the input $X$ the fast gradient sign method generates the adversarial images with : 
\begin{equation}\label{eq:fgsm_adv}
	X^{adv}=X + \epsilon \cdot sign(\bigtriangledown_{X}J(X,y_{true}))
\end{equation}
where $\epsilon$ is chosen to controls the perturbation's amplitude.

The Eqn.\ref{eq:fgsm_adv}
indicates that all pixels of the input $X$ are shifted simultaneously in the direction of the gradient with a single step. This method is simpler and faster than other methods, but lower attack success rate since at the beginning, it was designed to be fast rather than optimal.

\subsubsection{Iterative Gradient Sign}
An straightforward way introduced by Kurakin et al.\cite{kurakin2016adversarial} to extend the FGSM is to apply it several times with a smaller step size $\alpha$ and the intermediate result is clipped by the $\varepsilon$. Formally,
\begin{eqnarray}\label{eq:itv_adv}
	\begin{aligned}
		X_{0}^{adv} = &X, \\
		X_{N+1}^{adv} = &Clip_{X,\epsilon} \{X_{N}^{adv} + \\
		&\alpha \cdot sign(\bigtriangledown_{X}J(X_{N}^{adv},y_{true})) \}
	\end{aligned}
\end{eqnarray}

\subsubsection{DeepFool Attack}

DeepFool is a non-targeted attack introduced by Moosavi-Dezfooli et al \cite{moosavi2016deepfool}. It is one of methods to apply the minimal perturbation for misclassification under the $L_{2}$ distance metric. The method performs iterative steps on the adversarial direction of the gradient provided by a locally linear approximation of the classifier until the decision hyperplane has crossed. The objective of DeepFool is

\begin{eqnarray}\label{eq:DeepFool_adv}
	\begin{aligned}
		&minimize \quad ||D(x) - x||_2 \\
		&subject \ to  \quad \mathop{argmax} \limits_{y_{true}} \ f(D(x)) \neq y_{true} 
	\end{aligned}
\end{eqnarray}

Although it is different from L-BFGS, the attack can also be seen as a first-order method to craft adversarial perturbation.

\subsubsection{Jacobian-Based Saliency Map Attack(JSMA)}

The targeted attack is also a gradient-based method which uses the gradient to compute a saliency score for each pixel. The saliency score reflects how strongly each pixel can affect the resulting classification. Given the saliency map computed  by the model's Jacobian matrix, the attack greedily modifies the most important pixel at each iteration until the prediction has changed to a target class. This attack seeks to craft adversarial perturbation under $L_{0}$ distance metric \cite{papernot2016limitations}.

\subsubsection{Carlini and Wagner Attack(CW)}

There are three attacks for the $L_{0}$, $L_{2}$ and $L_{\infty}$ distance metric proposed by Carlini et al.\cite{carlini2017towards} Here, we just give a brief description of $L_{2}$ attack. The objective of $L_{2}$ attack is

\begin{eqnarray}\label{eq:l2_adv}
	\begin{aligned}
		minimize & \ ||\frac{1}{2}(tanh(\omega) + 1) - x||_2^2 \\ 
		& + c \cdot l(\frac{1}{2}(tanh(\omega) + 1)
	\end{aligned}
\end{eqnarray}
where the loss function $l$ is defined as 

\begin{small}
	\[\label{eq:f_adv}
	l(D(x)) = max(max\{Z(D(x))_i:i \neq t\} - Z(D(x))_t,-\kappa)
	\]
\end{small}

The $Z$ is the logits of a given model and $\kappa$ is used to control the confidence of adversarial examples. As $\kappa$ increases, the more powerful adversarial examples become. The constant c can be chosen with binary search which is similar to $\lambda$ in the L-BFGS attack. 

The Carlini and Wagner's attack is proved by the authors that it is superior to other published attacks. Then, all the three attacks should be taken into consideration to defense against.

In addition, we use CW-$L_0$, CW-$L_2$, CW-$L_\infty$ to represent the attack for the $L_{0}$, $L_{2}$ and $L_{\infty}$ distance metric respectively in the following experiments.

\subsection{Generative Adversarial Nets}

Generative Adversarial Net (GAN) is a framework incorporating an adversarial discriminator into the procedure of training generative models.
There are two models in the GAN: a generator $G$ that is optimized to estimate the data distribution and a discriminator $D$ that aims to distinguish between samples from the training data and fake samples from $G$.


The objective of GAN can be formulized as a minimax value function $V(G,D)$:
\begin{eqnarray}\label{gan_loss}
	\begin{aligned}
		\min_{G} \max_{D}  V(G,D) = &E_{X\sim p_{data}(X)}[\log D(X)] \\
		&+E_{z\sim p_{z}(z)}[\log (1-D(G(z)))]
	\end{aligned}
\end{eqnarray}

GAN has been known to be unstable to train, often resulting in generators that produce nonsensical outputs since it is difficult to maintain a balance between the G and D.


The Deep Convolutional Generative Adversarial Nets (DCGAN)\cite{radford2015unsupervised} is a good implementation of the GAN with convolutional networks that make them stable to train in most settings.

Our model in this paper is implemented based on DCGAN owing to its stability. The details will be discussed in the following section.

\section{Our Approach}
The fundamental idea of defending against adversarial examples is to eliminate or damage of the trivial perturbations of the input before being recognized by the target model.

The infinitesimal difference of adversarial image and clean image can be formulated as:
\begin{equation}
	\| X^{adv} - X\| = \eta
\end{equation}
Ideally, the perturbations $\eta$ can be got rid of from $X^{adv}$. That means the distribution of $X^{adv}$ is highly consistent with $X$.

The global optimality of GAN is the consistence of the generative distribution of $G$ with samples from the data generating distribution:
\begin{equation}
	p_{g} = p_{\textbf{data}}
\end{equation}
The procedure of converging to a good estimator of $p_{\textbf{data}}$ coincides with the demand of the elimimation of adversarial perturbations $\eta$.

Based on the above analysis, a novel framework based GAN to eliminate the adversarial perturbations 
is proposed. We name this class of architectures defending against adversarial examples based on GAN, adversarial perturbation elimination with GAN(APE-GAN),  as shown in Figure \ref{fig:generator}.

The APE-GAN  network is trained in an adversarial setting.
While the generator G is trained to alter the perturbation with tiny changes to the input examples, the discriminator $D$ is optimized to seperate the clean examples and reconstructed examples without adversarial perturbations obtained from G.
To achieve this, a task specified fusion loss function is invented to make the adversarial examples highly consistent with original clean image manifold.

\subsection{Architecture}
The ultimate goal of APE-GAN is to train a generating function $G$ that estimates for a given adversarial input image $X^{adv}$ its corresponding $\hat{X}$ counterpart.
To achieve this, a generator network parametrized by $\theta_{G}$ is trained. Here $\theta_{G}$ denotes the weights and baises of a generate network and is obtained by optimizing an adversarial perturbation elimination specified loss function $l_{ape}$. For training images $X_{k}^{adv}$ obtained by applying FGSM with corresponding original clean image $X_{k}$, $k=1,...,N$, we solve:
\begin{equation}
	\hat{\theta}_{G} = \arg \min_{\theta_{G}} \dfrac{1}{N} \sum_{k=1}^{N} l_{ape}(G_{\theta_{G}}(X_{k}^{adv}),X_{k})
\end{equation}

A discriminator network $D_{\theta_{D}}$ along with $G_{\theta_{G}}$ is defined to solve the adversarial zero sum problem:
\begin{eqnarray}\label{eq:disriminator}
	\begin{aligned}
		\min_{\theta_{G}} \max_{\theta_{D}} \quad  &\mathbb{E}_{X\sim p_{data}(X)}\log D_{\theta_{D}}(X) -\\
		&\mathbb{E}_{X^{adv}\sim p_{G}(X^{adv})}\log ( D_{\theta_{D}}(G_{\theta_{G}}(X^{adv})))
	\end{aligned}
\end{eqnarray}

\begin{table*}[tp]
	\begin{center}
		\fontsize{9}{10}\selectfont
		\begin{threeparttable}
			\caption{Error rates (in \%) of adversarial examples generated from five methods for target model and APE-GAN on MNIST, CIFAR10 and Imagenet. The error rates of target models on the clean images are reported in the experimental setup.}
			\label{tab:errorRate}
			\begin{tabular}{p{2.4cm}<{\centering} p{1.5cm}<{\centering} p{1.5cm}<{\centering} p{1.5cm}<{\centering} p{1.5cm}<{\centering} p{1.5cm}<{\centering} p{1.5cm}<{\centering}}
				\toprule
				\multirow{2}{*}{\textbf{Attack}}&
				\multicolumn{2}{c}{\textbf{MNIST}}&\multicolumn{2}{c}{\textbf{CIFAR10}} & \multicolumn{2}{c}{\textbf{ImageNet (Top-1)}}\cr
				\cmidrule(lr){2-3} \cmidrule(lr){4-5} \cmidrule(lr){6-7}
				&Target Model&APE-GAN$_m$ &Target Model&APE-GAN$_c$ &Target Model&APE-GAN$_i$\cr
				\midrule
				\textbf{L-BFGS}&93.4 &2.2 &92.7 &19.9&93.3 &42.9 \cr
				\textbf{FGSM} &96.3 & 2.8 &77.8 &26.4 &72.9 & 40.1  \cr
				\textbf{DeepFool}&97.1 & 2.2 &98.3 &19.2 &98.4 & 45.9  \cr
				\textbf{JSMA} &97.8 & 38.6 &94.1 &38.3 &98.7 & 45.0  \cr
				\textbf{CW-$L_0$} & 100.0 & 27.0 & 100.0 &46.9 & 100.0 & 29.4   \cr
				\textbf{CW-$L_2$} & 100.0 & 1.5 &100.0 &30.5 & 99.7 & 26.1  \cr
				\textbf{CW-$L_\infty$} & 100.0 & 1.2 &100.0 &32.2 & 100.0 & 27.0  \cr
				\bottomrule
			\end{tabular}
		\end{threeparttable}
	\end{center}
\end{table*}

The general idea behind this formulation is that it allows one to train a generative model $G$ with the goal of deceiting a differentiable discriminator $D$ that is trained to tell apart reconstructed images G($X^{adv}$) from original clean images. Consequently, the generator can be trained to produce reconstructed images that are highly similar to original clean images, and thus $D$ is unable to distinguish them.

The general architecture of our generator network $G$ is illustrated in Figure~\ref{fig:generator}. Some convolutional layers with stride = 2 are leveraged to get feature maps with lower resolution and followed by some deconvolutional layers with stride = 2 to recover the original resolution.

To discriminate original clean images from reconstructed images, we train a discriminator network. The general architecture is illustrated in Figure~\ref{fig:generator}. The discriminator network is trained to solve the maximization problem in Equation~\ref{eq:disriminator}. It also contains some convolutional layers with stride = 2 to get some high-level feature maps, two dense layers and a final sigmoid activation function to obtain a probability for sample classification.

The specific architectures on MNIST,CIFAR10 and ImageNet are introduced in
the experimental setup.

\subsection{Loss Function}

\subsubsection{Discriminator Loss}
According to Equation~\ref{eq:disriminator}, the loss function of discriminator , \textbf{$l_d$} is designed easily: 

\begin{eqnarray}\label{eq:dloss}
	\begin{aligned}
		l_d = \quad -&\sum_{n=1}^{N}[\log D_{\theta_{D}}(X) + \log D_{\theta_{D}}(G_{\theta_{G}}(X^{adv}))]
	\end{aligned}
\end{eqnarray}

\subsubsection{Generator Loss}
The definition of our adversarial perturbation elimination specified loss function $l_{ape}$ is critical for the performance of our generator network to produce images without adversarial perturbations. We define $l_{ape}$ as the weighted sum of several loss functions as:
\begin{equation} \label{eq:g_loss}
	l_{ape} =\xi_{1} l_{mse} + \xi_{2} l_{adv}
\end{equation}
which consists of pixel-wise MSE(mean square error) loss and adversarial loss.

\begin{itemize}
	\item \textbf{Content Loss:}
	Inspired by image super-resolution method\cite{ledig2016photo}, the pixel-wise MSE loss is defined as:
	\begin{equation}
		l_{mse} =  \dfrac{1}{WH}\sum_{x=1}^{W}\sum_{y=1}^{H}(X_{x,y}-G_{\theta_{G}}(X^{adv})_{x,y})^2
	\end{equation}
	Adversarial perturbations can be viewed as a special noise constructed delicately. The most widely used loss for image denoising or super-resolution will be able to achieve satisfactory results for adversarial perturbation elimination.
	
	\item \textbf{Adversarial Loss:} To encourage our network to produce images residing on the manifold of original clean images, the adversarial loss of the GAN is also employed. The adversarial loss $l_{adv}$ is calculated based on the probabilities of the discriminator over all reconstructed images:
	\begin{equation}
		l_{adv}=\sum_{n=1}^{N}[1-\log D_{\theta_{D}}(G_{\theta_{G}}(X^{adv}))]
	\end{equation}
	
\end{itemize}

\section{Evaluation}
The L-BFGS, DeepFool, JSMA, FGSM, CW including CW-L$_0$ CW-L$_2$ CW-L$_\infty$ attacks introduced in the related-work are resisted by APE-GAN on three standard datasets: MNIST\cite{lecun1998gradient}, a database of handwritten digits has 70,000 28x28 gray images in 10 classes(digit 0-9), CIFAR10\cite{krizhevsky2009learning}, a dataset consists of 60,000 32x32 colour images in 10 classes, and ImageNet\cite{deng2009imagenet}, a large-image recognition task with 1000 classes and more than 1,000,000 images provided.

It is noteworthy that the adversarial samples cannot be saved in the form of picture, since discretizing the values from a real-numbered value to one of the 256 points seriously degrades the quality. Then it should be saved and loaded as float32.

\begin{table*}[tp]
	\begin{center}
		\fontsize{9}{10}\selectfont
		\begin{threeparttable}
			\caption{Error rates (in \%) of benign input for target models and APE-GAN on MNIST, CIFAR10 and Imagenet. Here, the target models are model C, DenseNet40, InceptionV3 which are identical to the target models for FGSM attack on MNIST, CIFAR10 and Imagenet respectively.}
			\label{tab:srcError}
			\begin{tabular}{p{4.4cm}<{\centering} p{1.5cm}<{\centering} p{1.5cm}<{\centering} p{1.5cm}<{\centering} p{1.5cm}<{\centering} p{1.5cm}<{\centering} p{1.5cm}<{\centering}}
				\toprule
				\multirow{2}{*}{\textbf{Input}}&
				\multicolumn{2}{c}{\textbf{MNIST}}&\multicolumn{2}{c}{\textbf{CIFAR10}} & \multicolumn{2}{c}{\textbf{ImageNet (Top-1)}}\cr
				\cmidrule(lr){2-3} \cmidrule(lr){4-5} \cmidrule(lr){6-7}
				&C&APE-GAN$_m$ &DenseNet40&APE-GAN$_c$ &InceptionV3&APE-GAN$_i$\cr
				\midrule
				\textbf{clean image} &0.8 &1.2 &9.9 &10.3&22.9 &24.0 \cr
				\textbf{random Gaussian noise image} &1.7 &1.6 &11.3 &10.7 &25.2 & 24.5   \cr
				\bottomrule
			\end{tabular}
		\end{threeparttable}
	\end{center}
\end{table*}

\begin{table}[tp]
	\begin{center}
		\fontsize{9}{10}\selectfont
		\begin{threeparttable}
			\caption{Error rates (in \%) of adversarial exmamples generated from FGSM with different $\epsilon$  for target models, \emph{APE-GAN} on MNIST and CIFAR10. The error rates of target models on the clean images are reported in the experimental setup. Here, the target models are model C, DenseNet40 which are identical to the target models for FGSM attack on MNIST, CIFAR10 respectively.}
			\label{tab:cifarMnsitError}
			\begin{tabular}{p{1.2cm}<{\centering} p{1.2cm}<{\centering}  p{1.2cm}<{\centering} p{1.2cm}<{\centering} p{1.2cm}<{\centering}}
				\toprule
				\multirow{2}{*}{\textbf{Attack}}&
				\multicolumn{2}{c}{\textbf{MNIST}}&\multicolumn{2}{c}{\textbf{CIFAR10}} \cr
				\cmidrule(lr){2-3} \cmidrule(lr){4-5}
				&C&APE-GAN$_m$&DenseNet40&APE-GAN$_c$\cr
				\midrule
				$\epsilon$ = 0.1 &35.9 &0.8 &77.8 & 26.4\cr
				$\epsilon$ = 0.2 &86.0 &1.1 &84.7&45.2 \cr
				$\epsilon$ = 0.3 &96.3 &2.8 &86.3 &55.9  \cr
				$\epsilon$ = 0.4 &98.0 &21.0 & 87.2 & 63.4 \cr
				\bottomrule
			\end{tabular}
		\end{threeparttable}
	\end{center}
\end{table}

\subsection{Experimental Setup}
\subsubsection{Input} 
The input samples of target model can be classified into \emph{adversarial input} obtained from attack approaches and \emph{benign input} which is taken into account by the traditional deep learning framework including clean images and clean images added with random noise. Adding random noise to original clean images is the common trick used in data augmentation to improve the robustness, but does not belong to the standard training procedures of target model. Hence, it is not shown in Figure \ref{fig:generator}.

\begin{itemize}
	\item  \textbf{Adversarial Input:} The FGSM and JSMA attacks have been implemented in the \emph{cleverhans} v.1 \cite{papernot2016cleverhans} which is a Python library to benchmark machine learning systems' vulnerability to adversarial examples and the L-BFGS attack and DeepFool attacks have been implemented in the Foolbox\cite{rauber2017foolbox} which is a Python toolbox to create adversarial examples that fool neural networks. The code of CW attack has been provided by Carlini et al \cite{carlini2017towards}.
	We experiment with $\epsilon = 0.3$ on MNIST, $\epsilon = 0.1$ on CIFAE10, $\epsilon = 8 / 255$ on ImageNet for the FGSM attack, $\kappa$ = 0 on the three datasets for the CW-$L_2$ attack and default parameters of other attacks are utilised.
	\item  \textbf{Benign Input:}  The full test set of MNIST and CIFAR10 are utilized for the evaluation while results on ImageNet use a random sample of 10,000 RGB inputs from the test set. In addition, Gaussian white noise of mean 0 and variance 0.05 is employed in the following. 
\end{itemize}

\subsubsection{Target Models}
In order to provide the most accurate and fair comparison, whenever possible, the models provided by the authors or libraries should be used.

\begin{itemize}
	\item \textbf{MNIST:} We train a convolutional networks (denoted A in the Appendix) for L-BFGS and DeepFool attacks. For CW attack, the model is provided by Carlini (denoted B in the Appendix). For FGSM and JSMA attacks, the model is provided by \emph{cleverhans} (denoted C in the Appendix). The 0.9\%, 0.5\% and 0.8\% error rates can be achieved by the models A, B and C on clean images respectively, comparable to the state of the art.

	\item \textbf{CIFAR10:} \emph{ResNet18}\cite{he2016deep} is trained by us for L-BFGS and DeepFool attack. For CW attack, the model is provided by Carlini (denoted D in the Appendix). For FGSM and JSMA attacks, \emph{DenseNet}\cite{huang2017densely} with depth=40 is trained. The 7.1\%, 20.2\%\footnotemark[1] and 9.9\% error rates can be achieved by the models \emph{ResNet18}, D and \emph{DenseNet40}  on clean images respectively. 
	
	\footnotetext[1]{It is noteworthy that the 20.2\% error rate of target model D is significantly greater than the other models, however, identical to the error rate reported by Carlini\cite{carlini2017towards}. For accurate comparison, we respect the choice of author.}
	
	\item \textbf{ImageNet:} We use \emph{ResNet50} one pre-trained networks for L-BFGS and DeepFool attacks. For other three attacks, another pre-trained network \emph{InceptionV3} is leveraged \cite{russakovsky2015imagenet}. \emph{ResNet50} achieves the top-1 error rate 24.4\% and the top-5 error rate 7.2\% while \emph{InceptionV3} achieves the top-1 error rate 22.9\%  and the top-5 error rate  6.1\% on clean images.
\end{itemize}

\begin{table}[tp]
	\begin{center}
		\fontsize{9}{10}\selectfont
		\begin{threeparttable}
			\caption{Error rates (in \%) of adversarial exmamples generated from FGSM with different $\epsilon$  for target model InceptionV3, \emph{APE-GAN$_i$} on ImageNet. The error rate of target model on the clean images is reported in the experimental setup.}
			\label{tab:imageNetError}
			\begin{tabular}{p{1.5cm}<{\centering} p{1.1cm}<{\centering}  p{1.2cm}<{\centering} p{1.1cm}<{\centering}  p{1.2cm}<{\centering}}
				\toprule
				\multirow{2}{*}{\textbf{Attack}}&
				\multicolumn{2}{c}{\textbf{ImageNet(Top 1)}} & \multicolumn{2}{c}{\textbf{ImageNet(Top 5)}}\cr
				\cmidrule(lr){2-3} \cmidrule(lr){4-5}
				&InceptionV3&APE-GAN$_i$ &InceptionV3&APE-GAN$_i$\cr
				\midrule
				$\epsilon$ = 4 / 255 &72.2 &38.0 & 41.7 & 21.0\cr
				$\epsilon$ = 8 / 255 &72.9 &40.1 & 42.3 & 22.5\cr
				$\epsilon$ = 12 / 255 &73.4 &41.2& 43.4 & 22.9\cr
				$\epsilon$ = 16 / 255 &74.8 &42.4& 44.0 & 23.6\cr
				\bottomrule
			\end{tabular}
		\end{threeparttable}
	\end{center}
\end{table}

\subsubsection{APE-GAN}
Three models are trained with the APE-GAN architecture on MNIST, CIFAR10 and ImageNet(denoted APE-GAN$_m$, APE-GAN$_c$, APE-GAN$_i$ in the Appendix).
The full training set of MNIST and CIFAR10 are utilized for the training of APE-GAN$_m$ and APE-GAN$_c$ respectively while a random sample of 50,000 RGB inputs from the training set of ImageNet make a contribution to train the APE-GAN$_i$.

\begin{itemize}
	\item \textbf{MNIST:} APE-GAN$_m$ is trained for 2 epochs on batches of 64 FGSM samples with $\epsilon$ = 0.3, input size = (28,28,1). 
	
	\item \textbf{CIFAR10:} APE-GAN$_c$ is trained for 10 epochs on batches of 64 FGSM samples with $\epsilon$ = 0.1, input size = (32,32,3). 
	
	\item \textbf{ImageNet:} APE-GAN$_i$ is trained for 30 epochs on batches of 16 FGSM samples with $\epsilon$ = 8 / 255, input size = (256,256,3). However, as we all know, the input size of ResNet50 is 224 * 224 and the InceptionV3 is 299 * 299. So we use the resize operation to handle this. 
\end{itemize}

The straightforward method to train the generator and the
discriminator is update both in every batch. However, the
discriminator network often learns much faster than the
generator network because the generator is more complex
than distinguishing between real samples and fake samples. Therefore, generator
should be run twice with each iteration to make sure that
the loss of discriminator does not go to zero. 
The learning rate is initialized with 0.0002 and Adam\cite{kingma2014adam} optimizer is used to update parameters and optimize the networks. The weights of the adversarial perturbation elimination specified loss $\xi_1$ and $\xi_2$ used in the Eqn.\ref{eq:g_loss} are fixed to 0.7 and 0.3 separately.

The training procedure of the APE-GAN
needs no knowledge of the architecture and parameters of the target model.

\subsection{Results}

\subsubsection{Effectiveness:} 
\begin{figure}
	\centering
	\begin{center}
		\subfigure[]{\includegraphics[width=0.45\linewidth]{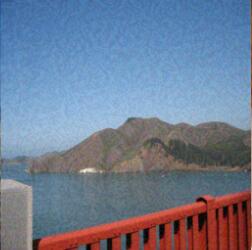}}
		\subfigure[]{\includegraphics[width=0.45\linewidth]{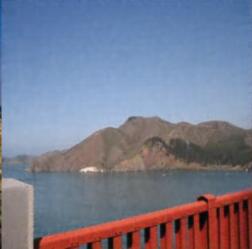}}
	\end{center}
	\caption{ImageNet dataset
		(a) Adversarial samples crafted by FGSM  with $\epsilon$ = 8 / 255 on the ImageNet   (b) Reconstructed samples by APE-GAN }
	\label{fig:imageNet_denoise}
\end{figure}

\begin{figure}
	\centering
	\begin{center}
		\subfigure[]{\includegraphics[width=0.3\linewidth]{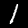}}
		\subfigure[]{\includegraphics[width=0.3\linewidth]{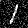}}
		\subfigure[]{\includegraphics[width=0.3\linewidth]{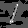}}
		\subfigure[]{\includegraphics[width=0.3\linewidth]{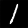}}
		\subfigure[]{\includegraphics[width=0.3\linewidth]{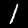}}
		\subfigure[]{\includegraphics[width=0.3\linewidth]{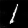}}
	\end{center}
	\caption{MNIST dataset \textbf{(a)}. clean image \textbf{(b)}. random Gaussian noise image \textbf{(c)}. adversarial samples obtained from FGSM with $\epsilon$ = 0.3 \textbf{(d)}. reconstructed image of (a) by APE-GAN \textbf{(e)}. reconstructed image of (b) by APE-GAN  \textbf{(f)}. reconstructed samples of (c) by APE-GAN }
	\label{fig:src_noise_mnist}
\end{figure}

\begin{itemize}
	\item \textbf{Adversarial Input} Table \ref{tab:errorRate} indicates that the error rates of adversarial inputs are significantly decreased after its perturbation is eliminated by APE-GAN. Among all the attacks, the CW attack is more offensive than the others, and among the three attacks of CW, the CW-$L_0$ is more offensive. The error rate of FGSM is greater than the L-BFGS which may be caused by different target models.  As it is shown in Figure \ref{fig:imageNet_denoise}, the aggressivity of adversarial examples can be eliminated by APE-GAN even though these is imperceptible differences between (a) and (b). 
	In addition, the adversarial examples generated from FGSM with different  $\epsilon$ are resisted and the result is shown in Table \ref{tab:cifarMnsitError} and Table \ref{tab:imageNetError}.
	
	\item \textbf{Benign Input}
	The error rate of clean images and the clean images added with random Gaussian noise is shown in Table~\ref{tab:srcError}.	
	Actual details within the image can be lost with multiple levels of convolutional and down-sampling layers which has a negative effect on the classification. However, Table~\ref{tab:srcError} indicates that there is no marked increase in the error rate of clean images. Meanwhile, APE-GAN has a good performance on resisting the random noise. Figure \ref{fig:src_noise_mnist} shows that the perturbation generated from random Gaussian noise is irregular and all in a muddle while the perturbation obtained from the FGSM attack is regular and intentional.
	However the perturbation, whether regular or irregular, can be eliminated by APE-GAN.
	
\end{itemize}

In summary, APE-GAN has the capability to provide a good performance to  various input, whether adversarial or benign, on three benchmark datasets.



\subsubsection{Strong Applicability:} The experimental setup of target models indicates that there is more than one target model designed in experiments on MNIST, CIFAR10 and ImageNet respectively. Table \ref{tab:errorRate} demonstrates that APE-GAN can tackle adversarial examples for different target models.
Actually, it can provide a defense without knowing what model they are constructed upon. Therefore, we can conclude that the APE-GAN possesses strong applicability.

\section{Discussion and Future Work}

Pre-processing the input to eliminate the adversarial perturbations is another appealing aspect of the framework which makes sure there is no conflict between the framework and other existing defenses.
Then \textbf{APE-GAN can work with other defenses such as adversarial training together}. Another method APE-GAN followed by a target model trained using adversarial training is experimented. The results on MNIST and CIFAR10 have been done shown in Table {\color{red}{7}}, {\color{red}{8}}, {\color{red}{9}} in the Appendix. The adversarial examples leveraged in Table {\color{red}{9}} in the Appendix are generated from \emph{Iterative Gradient Sign} with N = 2. Actually, the \emph{FGSM} leveraged to craft the adversarial examples of Table {\color{red}{8}} is identical to the \emph{Iterative Gradient Sign} with N = 1. Compared with Table {\color{red}{8}}, Table {\color{red}{9}} indicates that the robustness of target model cannot be significantly improved with adversarial training. However, the combination of APE-GAN and adversarial training makes a notable defence against \emph{Iterative Gradient Sign}. New combinations of different defenses will be researched in the future work.


The core work in this paper is to propose a new perspective of defending against adversarial examples and to first eliminate the adversarial perturbations using a trained network and then feed the processed example to classification networks. The training of this adversarial perturbation elimination network is based on the Generative Adversarial Nets framework.
Experimental results on three benchmark datasets demonstrate the effectiveness of the proposed approach.

{\small
\bibliographystyle{ieee}
\bibliography{egbib}

\begin{thebibliography}{10}\itemsep=-1pt

\bibitem{carlini2017adversarial}
N.~Carlini and D.~Wagner.
\newblock Adversarial examples are not easily detected: Bypassing ten detection
  methods.
\newblock {\em arXiv preprint arXiv:1705.07263}, 2017.

\bibitem{carlini2017towards}
N.~Carlini and D.~Wagner.
\newblock Towards evaluating the robustness of neural networks.
\newblock In {\em Security and Privacy (SP), 2017 IEEE Symposium on}, pages
  39--57. IEEE, 2017.

\bibitem{deng2009imagenet}
J.~Deng, W.~Dong, R.~Socher, L.-J. Li, K.~Li, and L.~Fei-Fei.
\newblock Imagenet: A large-scale hierarchical image database.
\newblock In {\em Computer Vision and Pattern Recognition, 2009. CVPR 2009.
  IEEE Conference on}, pages 248--255. IEEE, 2009.

\bibitem{gong2017adversarial}
Z.~Gong, W.~Wang, and W.-S. Ku.
\newblock Adversarial and clean data are not twins.
\newblock {\em arXiv preprint arXiv:1704.04960}, 2017.

\bibitem{goodfellow2014generative}
I.~Goodfellow, J.~Pouget-Abadie, M.~Mirza, B.~Xu, D.~Warde-Farley, S.~Ozair,
  A.~Courville, and Y.~Bengio.
\newblock Generative adversarial nets.
\newblock In {\em Advances in neural information processing systems}, pages
  2672--2680, 2014.

\bibitem{goodfellow2014explaining}
I.~J. Goodfellow, J.~Shlens, and C.~Szegedy.
\newblock Explaining and harnessing adversarial examples.
\newblock {\em arXiv preprint arXiv:1412.6572}, 2014.

\bibitem{he2016deep}
K.~He, X.~Zhang, S.~Ren, and J.~Sun.
\newblock Deep residual learning for image recognition.
\newblock In {\em Proceedings of the IEEE conference on computer vision and
  pattern recognition}, pages 770--778, 2016.

\bibitem{huang2017densely}
G.~Huang, Z.~Liu, L.~van~der Maaten, and K.~Q. Weinberger.
\newblock Densely connected convolutional networks.
\newblock In {\em Proceedings of the IEEE Conference on Computer Vision and
  Pattern Recognition}, 2017.

\bibitem{kingma2014adam}
D.~Kingma and J.~Ba.
\newblock Adam: A method for stochastic optimization.
\newblock {\em arXiv preprint arXiv:1412.6980}, 2014.

\bibitem{krizhevsky2009learning}
A.~Krizhevsky and G.~Hinton.
\newblock Learning multiple layers of features from tiny images.
\newblock 2009.

\bibitem{kurakin2016adversarial}
A.~Kurakin, I.~Goodfellow, and S.~Bengio.
\newblock Adversarial examples in the physical world.
\newblock {\em arXiv preprint arXiv:1607.02533}, 2016.

\bibitem{lecun1998gradient}
Y.~LeCun, L.~Bottou, Y.~Bengio, and P.~Haffner.
\newblock Gradient-based learning applied to document recognition.
\newblock {\em Proceedings of the IEEE}, 86(11):2278--2324, 1998.

\bibitem{ledig2016photo}
C.~Ledig, L.~Theis, F.~Husz{\'a}r, J.~Caballero, A.~Cunningham, A.~Acosta,
  A.~Aitken, A.~Tejani, J.~Totz, Z.~Wang, et~al.
\newblock Photo-realistic single image super-resolution using a generative
  adversarial network.
\newblock {\em arXiv preprint arXiv:1609.04802}, 2016.

\bibitem{liu2016delving}
Y.~Liu, X.~Chen, C.~Liu, and D.~Song.
\newblock Delving into transferable adversarial examples and black-box attacks.
\newblock {\em arXiv preprint arXiv:1611.02770}, 2016.

\bibitem{metzen2017detecting}
J.~H. Metzen, T.~Genewein, V.~Fischer, and B.~Bischoff.
\newblock On detecting adversarial perturbations.
\newblock {\em arXiv preprint arXiv:1702.04267}, 2017.

\bibitem{moosavi2016deepfool}
S.-M. Moosavi-Dezfooli, A.~Fawzi, and P.~Frossard.
\newblock Deepfool: a simple and accurate method to fool deep neural networks.
\newblock In {\em Proceedings of the IEEE Conference on Computer Vision and
  Pattern Recognition}, pages 2574--2582, 2016.

\bibitem{papernot2016cleverhans}
N.~Papernot, I.~Goodfellow, R.~Sheatsley, R.~Feinman, and P.~McDaniel.
\newblock cleverhans v1.0.0: an adversarial machine learning library.
\newblock {\em arXiv preprint arXiv:1610.00768}, 2016.

\bibitem{papernot2017extending}
N.~Papernot and P.~McDaniel.
\newblock Extending defensive distillation.
\newblock {\em arXiv preprint arXiv:1705.05264}, 2017.

\bibitem{papernot2016transferability}
N.~Papernot, P.~McDaniel, and I.~Goodfellow.
\newblock Transferability in machine learning: from phenomena to black-box
  attacks using adversarial samples.
\newblock {\em arXiv preprint arXiv:1605.07277}, 2016.

\bibitem{papernot2016practical}
N.~Papernot, P.~McDaniel, I.~Goodfellow, S.~Jha, Z.~B. Celik, and A.~Swami.
\newblock Practical black-box attacks against deep learning systems using
  adversarial examples.
\newblock {\em arXiv preprint arXiv:1602.02697}, 2016.

\bibitem{papernot2016limitations}
N.~Papernot, P.~McDaniel, S.~Jha, M.~Fredrikson, Z.~B. Celik, and A.~Swami.
\newblock The limitations of deep learning in adversarial settings.
\newblock In {\em Security and Privacy (EuroS\&P), 2016 IEEE European Symposium
  on}, pages 372--387. IEEE, 2016.

\bibitem{papernot2016distillation}
N.~Papernot, P.~McDaniel, X.~Wu, S.~Jha, and A.~Swami.
\newblock Distillation as a defense to adversarial perturbations against deep
  neural networks.
\newblock In {\em Security and Privacy (SP), 2016 IEEE Symposium on}, pages
  582--597. IEEE, 2016.

\bibitem{radford2015unsupervised}
A.~Radford, L.~Metz, and S.~Chintala.
\newblock Unsupervised representation learning with deep convolutional
  generative adversarial networks.
\newblock {\em arXiv preprint arXiv:1511.06434}, 2015.

\bibitem{rauber2017foolbox}
J.~Rauber, W.~Brendel, and M.~Bethge.
\newblock Foolbox v0.8.0: A python toolbox to benchmark the robustness of
  machine learning models.
\newblock {\em arXiv preprint}, 2017.

\bibitem{russakovsky2015imagenet}
O.~Russakovsky, J.~Deng, H.~Su, J.~Krause, S.~Satheesh, S.~Ma, Z.~Huang,
  A.~Karpathy, A.~Khosla, M.~Bernstein, et~al.
\newblock Imagenet large scale visual recognition challenge.
\newblock {\em International Journal of Computer Vision}, 115(3):211--252,
  2015.

\bibitem{szegedy2013intriguing}
C.~Szegedy, W.~Zaremba, I.~Sutskever, J.~Bruna, D.~Erhan, I.~Goodfellow, and
  R.~Fergus.
\newblock Intriguing properties of neural networks.
\newblock {\em arXiv preprint arXiv:1312.6199}, 2013.

\bibitem{tramer2017ensemble}
F.~Tram{\`e}r, A.~Kurakin, N.~Papernot, D.~Boneh, and P.~McDaniel.
\newblock Ensemble adversarial training: Attacks and defenses.
\newblock {\em arXiv preprint arXiv:1705.07204}, 2017.

\end{thebibliography}
}

\clearpage

\begin{table*}[tp]
	\begin{center}
		\fontsize{9}{10}\selectfont
		\caption{\quad \textbf{Conv1}: convolutional layer with stride = 1. \textbf{Conv2}: convolutional layer with stride = 2. \textbf{FC}: fully connected layer. Model A,B,C are leveraged for MNIST and model D for CIFAR10.}
		\begin{tabular}{p{2.8cm}<{\centering}  p{2.8cm}<{\centering}  p{3.0cm}<{\centering} p{3.0cm}<{\centering}}
			\hline
			A&B&C&D\\
			\hline
			Conv1(32,3,3) + Relu&Conv1(32,3,3) + Relu &Conv2(64,8,8) + Relu  & Conv1(64,3,3) + Relu\\
			Conv1(64,3,3) + Relu & Conv1(32,3,3) + Relu & Conv2(128,6,6) + Relu   & Conv1(64,3,3) + Relu\\
			Max Pooling        &Max Pooling & Conv1(128,5,5) + Relu  &Max Pooling\\
			Dropout(0.25)     &Conv1(64,3,3) + Relu & FC(10) + Softmax &Conv1(128,3,3) + Relu \\
			FC(128) + Relu & Conv1(64,3,3) + Relu &  & Conv1(128,3,3) + Relu\\
			Dropout(0.5)  & Max Pooling &  & Max Pooling \\
			FC(10) + Softmax & FC(200) + Relu&    & FC(256) + Relu\\
			& FC(200) + Relu&   & FC(256) + Relu\\
			& FC(10) + Softmax&  & FC(10) + Softmax \\
			\hline
		\end{tabular}
	\end{center}
\end{table*}

\begin{table*}[tp]
	\begin{center}
		\fontsize{9}{10}\selectfont
		\begin{threeparttable}
			\caption{The following three models are trained based on the AE-GAN framework on MNIST, CIFAR10 and ImageNet. \textbf{Conv}: convolutional layer with stride = 2. \textbf{Deconv}: conv2d\_transpose layer with stride = 2. \textbf{FC}: fully connected layer. \textbf{BN}: Batch Normalization.  \textbf{C}: channel numbers of the input. 
				The input of generator is an image not the vector \textbf{\emph{Z}} which is the major difference from original GAN. The model APE-GAN$_m$ has no difference from APE-GAN$_c$ except for the input size and the channel numbers.}
			\label{tab:acc_advgtsrb}
			\begin{tabular}{p{3.2cm}<{\centering} p{3.2cm}<{\centering} p{3.2cm}<{\centering} p{3.2cm}<{\centering} }
				\toprule
				\multicolumn{2}{c}{APE-GAN$_m$ , APE-GAN$_c$}&\multicolumn{2}{c}{APE-GAN$_i$}\cr
				\cmidrule(lr){1-2} \cmidrule(lr){3-4}
				generator&discriminator&generator&discriminator\cr
				\midrule
				Conv(64,3,3)&Conv(64,3,3) + lrelu &Conv(64,3,3) &Conv(64,3,3) \cr
				BN + lrelu &Conv(128,3,3)  & BN + lrelu &BN + lrelu   \cr
				Conv(128,3,3)&BN + lrelu & Conv(128,3,3)&Conv(128,3,3)     \cr
				BN + lrelu&Conv(256,3,3)  & BN + lrelu &BN + lrelu \cr
				
				Deconv(64)&BN + lrelu & Conv(256,3,3) &Conv(256,3,3) \cr
				
				BN + lrelu & FC(1) + Sigmoid & BN + lrelu & BN + lrelu  \cr
				Deconv(\textbf{C})& & Deconv(256) & Conv(256,3,3)\cr
				
				Sigmoid & & BN + lrelu & BN + lrelu\cr
				
				 &  & Deconv(128) & FC(1024) \cr
				 &   & BN + lrelu & BN + lrelu \cr
				
				&  & Deconv(\textbf{C}) & FC(1) + Sigmoid \cr
				&   & Sigmoid &  \cr
				\bottomrule
			\end{tabular}
		\end{threeparttable}
	\end{center}
\end{table*}

\begin{table*}[tp]
	\begin{center}
		\fontsize{9}{10}\selectfont
		\begin{threeparttable}
			\caption{Error rates (in \%) of adversarial exmamples generated from five attack methods for \emph{target model}, \emph{AE-GAN}, \emph{adversarial training} and \emph{AE-GAN + adversarial training} on MNIST and CIFAR10. The error rates of target models on the clean images are reported in the experimental setup. 	
				The error rates of adversarial examples generated from L-BFGS(based on the target model ResNet18), DeepFool(based on the target model ResNet18) and CW (based on the target model D) are 10.7\%, 12.5\%, 11.8\% on CIFAR10 respectively. It 
				illustrates that the \emph{transferability} doesn't work when we utilize the adversarial exmamples generated from model ResNet18, D to attck model Desnet40. Then the error rates of \emph{adversarial training} and \emph{AE-GAN + adversarial training} are filled with default values.
			}
			
			\label{tab:at1_five}
			\begin{tabular}{p{1.8cm}<{\centering} p{1.4cm}<{\centering} p{1.4cm}<{\centering} p{1.4cm}<{\centering} p{1.4cm}<{\centering} p{1.4cm}<{\centering} p{1.4cm}<{\centering} p{1.4cm}<{\centering} p{1.4cm}<{\centering}}
				\toprule
				\multirow{2}{*}{\textbf{Attack}}&
				\multicolumn{4}{c}{\textbf{MNIST}}&\multicolumn{4}{c}{\textbf{CIFAR10}} \cr
				\cmidrule(lr){2-5} \cmidrule(lr){6-9}
				&Target Model&$AE-GAN_m$ &Adversarial Training&$AE-GAN_m$ + adversarial training &Target Model&$AE-GAN_c$ &Adversarial Training&$AE-GAN_c$ + adversarial training\cr
				\midrule
				\textbf{L-BFGS}&93.4 &2.2 &2.6 & \textbf{2.0}&92.7 &19.0 & - & -\cr
				\textbf{FGSM} &96.3 & 2.8 &6.6 & \textbf{2.6}&77.8 &26.4 & \textbf{12.2} & 23.7\cr
				\textbf{DeepFool}&97.1 & 2.2 &3.7&\textbf{2.1}&98.3 &19.2& - & - \cr
				\textbf{JSMA} &97.8 & 38.6 &33.6&\textbf{21.0} &94.1 &38.3 & 55.3 & \textbf{32.3} \cr
				\textbf{CW-$L_0$} & 100.0 & 27.0 &15.5& \textbf{9.5}& 100.0 & 46.9 & -  &-   \cr
				\textbf{CW-$L_2$} & 100.0 & 1.5 &2.5 & \textbf{1.1} &100.0 &30.5 & - & -  \cr
				\textbf{CW-$L_\infty$} & 100.0 & 1.2 & 0.7&\textbf{0.5} &100.0 &32.2 & - & -  \cr
				\bottomrule
			\end{tabular}
		\end{threeparttable}
	\end{center}
\end{table*}

\begin{table*}[tp]
	\begin{center}
		\fontsize{9}{10}\selectfont
		\begin{threeparttable}
			\caption{Error rates (in \%) of adversarial exmamples generated from FGSM with different $\epsilon$  for \emph{target model}, \emph{AE-GAN}, \emph{adversarial training} and \emph{AE-GAN + adversarial training} on MNIST and CIFAR10. The error rates of target models on the clean images are reported in the experimental setup. Here, the target models used for FGSM attack are also leveraged as the base models for \emph{adversarial training}.}
			\label{tab:at1_fgsm}
			\begin{tabular}{p{1.8cm}<{\centering} p{1.4cm}<{\centering} p{1.4cm}<{\centering} p{1.4cm}<{\centering} p{1.4cm}<{\centering} p{1.4cm}<{\centering} p{1.4cm}<{\centering} p{1.4cm}<{\centering} p{1.4cm}<{\centering}}
				\toprule
				\multirow{2}{*}{\textbf{Attack}}&
				\multicolumn{4}{c}{\textbf{MNIST}}&\multicolumn{4}{c}{\textbf{CIFAR10}} \cr
				\cmidrule(lr){2-5} \cmidrule(lr){6-9}
				&Target Model&$AE-GAN_m$ &Adversarial Training&$AE-GAN_m$ + adversarial training &Target Model&$AE-GAN_c$ &Adversarial Training&$AE-GAN_c$ + adversarial training\cr
				\midrule
				$\epsilon$ = 0.1 &35.9 &0.8 &1.6 & \textbf{0.6}&77.8 & 26.4 & \textbf{12.2} & 23.7\cr
				$\epsilon$ = 0.2 &86.0 &1.1 &3.4 & \textbf{1.0}&84.7&45.2 & 39.6 & \textbf{36.4}\cr
				$\epsilon$ = 0.3 &96.3 &2.8 &6.6&\textbf{2.6}&86.3 &55.9 & 73.7 & \textbf{53.7} \cr
				$\epsilon$ = 0.4 &98.0 &21.0 &59.8&\textbf{17.7} & 87.2 & 63.4 & 81.7 & \textbf{63.2}\cr
				\bottomrule
			\end{tabular}
		\end{threeparttable}
	\end{center}
\end{table*}


\begin{table*}[tp]
	\begin{center}
		\fontsize{9}{10}\selectfont
		\begin{threeparttable}
			\caption{Error rates (in \%) of adversarial exmamples generated from Iterative Gradient Sign with N = 2 and different $\alpha$ for \emph{adversarial training} and \emph{AE-GAN + adversarial training} on MNIST and CIFAR10. Here, the target models are used for Iterative Gradient Sign attack which are identical to the target models for FGSM attack.
			}
			\label{tab:at1_igsm}
			\begin{tabular}{p{1.8cm}<{\centering} p{1.8cm}<{\centering} p{1.8cm}<{\centering} p{1.8cm}<{\centering} p{1.8cm}<{\centering}}
				\toprule
				\multirow{2}{*}{\textbf{Attack}}&
				\multicolumn{2}{c}{\textbf{MNIST}}&\multicolumn{2}{c}{\textbf{CIFAR10}} \cr
				\cmidrule(lr){2-3} \cmidrule(lr){4-5}
				&Adversarial Training & $AE-GAN_m$ + adversarial training&Adversarial Training&$AE-GAN_c$ + adversarial training \cr
				\midrule
				$\alpha$ = 0.1 & 32.0 & 1.9 & 72.6 & 26.5\cr
				$\alpha$ = 0.2 & 82.5 & 2.5 & 82.5 & 38.3\cr
				$\alpha$ = 0.3 & 84.7 & 4.5 & 86.5 & 51.7\cr
				$\alpha$ = 0.4 & 94.5 & 22.3& 88.0 & 71.0\cr
				\bottomrule
			\end{tabular}
		\end{threeparttable}
	\end{center}
\end{table*}

\begin{figure*}
	\centering
	\begin{center}
		\subfigure[adversarial samples obtained from FGSM with $\epsilon$ = 8 / 255]{\includegraphics[width=0.48\linewidth]{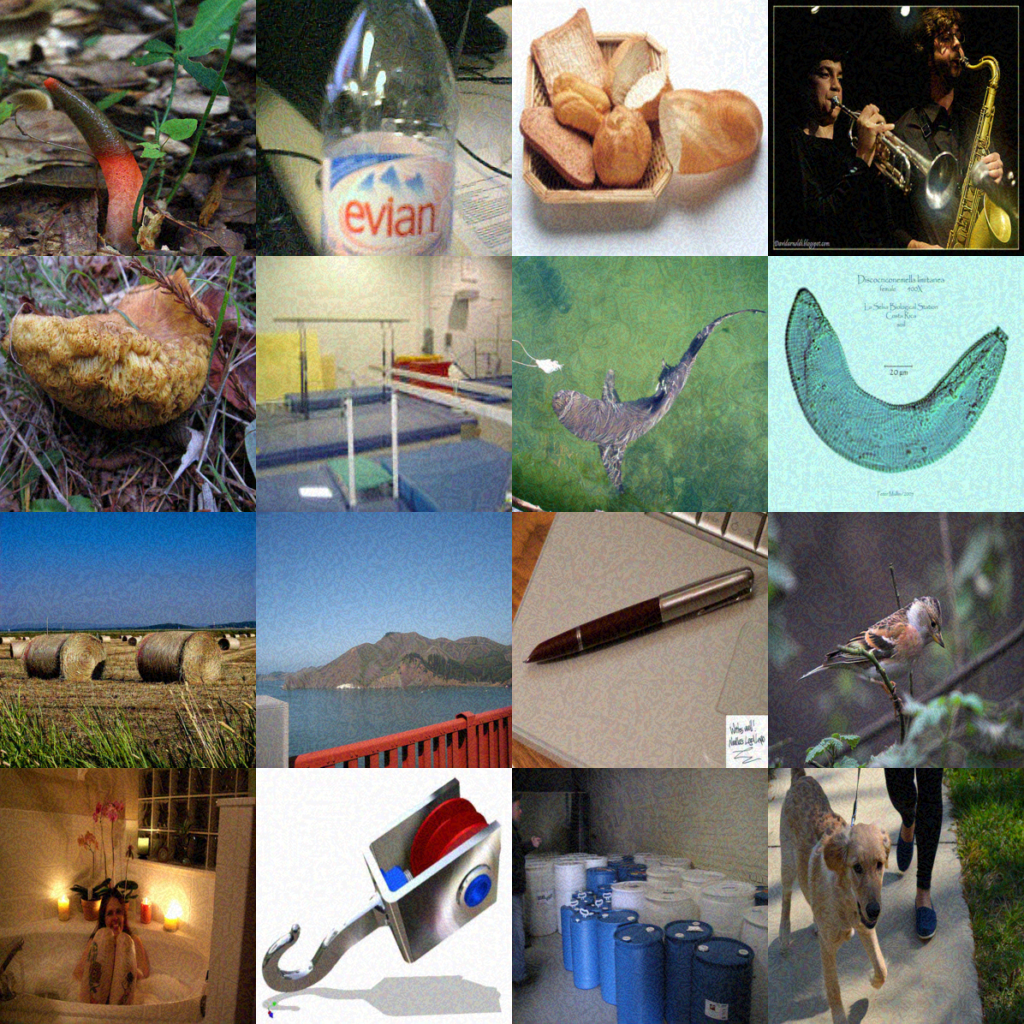}}
		\subfigure[reconstructed image by APE-GAN]{\includegraphics[width=0.48\linewidth]{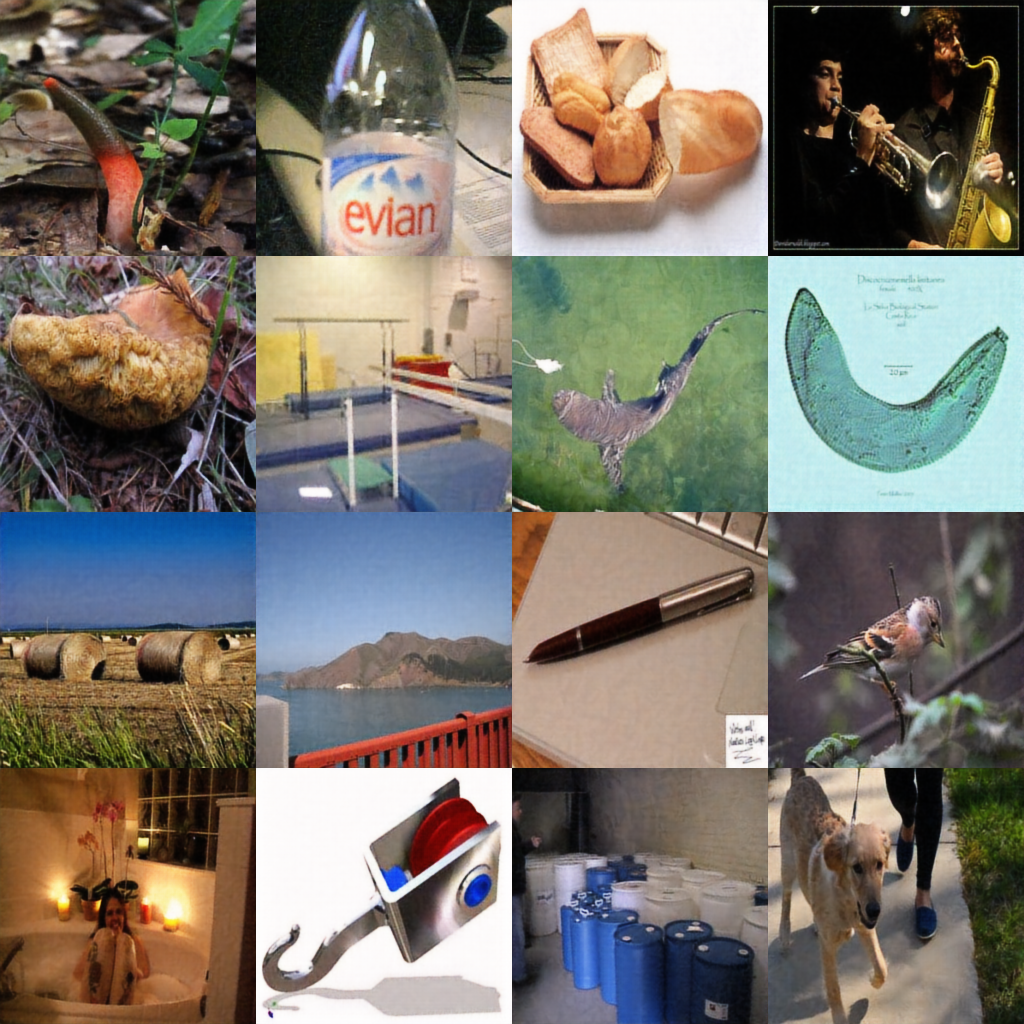}}
	\end{center}
	\caption{ImageNet dataset}. 
	\label{fig:ImageNet_res}
\end{figure*}

\end{document}